\documentclass{article}

\usepackage[preprint]{neurips_2025}

\usepackage[utf8]{inputenc} % allow utf-8 input
\usepackage[T1]{fontenc}    % use 8-bit T1 fonts
\usepackage{hyperref}       % hyperlinks
\usepackage{url}            % simple URL typesetting
\usepackage{booktabs}       % professional-quality tables
\usepackage{amsfonts}       % blackboard math symbols
\usepackage{nicefrac}       % compact symbols for 1/2, etc.
\usepackage{microtype}      % microtypography
\usepackage{xcolor}         % colors
\usepackage{graphicx}
\usepackage{subfigure}
\usepackage{subcaption}
\usepackage{multirow}
\usepackage{tabulary}
\usepackage{amsmath}
\usepackage[most]{tcolorbox}

\title{Advancing Brainwave Modeling with a Codebook-Based Foundation Model}

\author{
Konstantinos Barmpas\textsuperscript{*} \\
Imperial College London \& Cogitat \\
\texttt{konstantinos.barmpas16@imperial.ac.uk} \\
\And
Na Lee \textsuperscript{*} \\
Imperial College London \& Cogitat \\
\texttt{na.lee12@ic.ac.uk} \\
\And
Yannis Panagakis \\
National and Kapodistrian University of Athens \& \\Archimedes Research Unit \& Cogitat \\
\And
Dimitrios A. Adamos \\
Imperial College London \& Cogitat \\
\And 
Nikolaos Laskaris \\
Aristotle University of Thessaloniki \& Cogitat \\
\And
Stefanos Zafeiriou \\
Imperial College London \& Cogitat \\
}

\begin{document}

\maketitle

\renewcommand{\thefootnote}{\fnsymbol{footnote}} % Changes footnote style to symbols
\footnotetext[1]{These authors contributed equally to this work.}

\begin{abstract}
Recent advances in large-scale pre-trained Electroencephalogram (EEG) models have shown great promise, driving progress in Brain-Computer Interfaces (BCIs) and healthcare applications. However, despite their success, many existing pre-trained models have struggled to fully capture the rich information content of neural oscillations, a limitation that fundamentally constrains their performance and generalizability across diverse BCI tasks. This limitation is frequently rooted in suboptimal architectural design choices which constrain their representational capacity. In this work, we introduce LaBraM++, an enhanced Large Brainwave Foundation Model (LBM) that incorporates principled improvements grounded in robust signal processing foundations. LaBraM++ demonstrates substantial gains across a variety of tasks, consistently outperforming its originally-based architecture and achieving competitive results when compared to other open-source LBMs. Its superior performance and training efficiency highlight its potential as a strong foundation for future advancements in LBMs.
\end{abstract}

\section{Introduction}

Brain-Computer Interface (BCI) technology enables a direct communication of the brain with the external world. This is attainable by analyzing brainwaves captured by electroencephalogram (EEG) recorders using advanced signal processing and, more recently, machine learning techniques. The roots of brainwave signal analysis lie in traditional neuroscience, where lengthy studies and hand-crafted features were once considered the gold standard [\citet{SignalProcessingMethods1}, \citet{SignalProcessingMethods2}, \citet{SignalProcessingMethods3}, \citet{SignalProcessingMethods4}, \citet{Classical}]. However, as later research revealed, individual differences in brain activity (known as inter-subject variability) often prevent these features from generalizing effectively to real-world data, limiting their practicality in everyday BCI applications.

As we have transitioned into the era of data-driven solutions, deep learning has reduced the need for manual feature extraction. Deep models can now automatically extract relevant features for various BCI paradigms, often achieving impressive performance [\citet{EEGNet, EEGInception, EEGConformer, Barmpas_Scattering}]. Drawing inspiration from neuroscience \citet{Barmpas_2024_causality_jne}, machine learning researchers have integrated domain insights into their models, addressing challenges such as inter-subject variability [\citet{ourNeurISP2021, BEETL}]. However, despite their successes, deep learning models typically require significant supervision and task-specific data collection, making the development process time consuming and resource-intensive.

More recently, Large Foundation Models have transformed fields such as Computer Vision and Natural Language Processing with their remarkable results  [\citet{brown2020language, touvron2023llama, mizrahi2023m, paraperas2024arc2face}]. These large models offer several advantages, including the ability to capture complex patterns through extensive self-supervised pre-training on diverse, unlabeled datasets. Therefore, they exhibit strong generalization capabilities, reducing the need for task-specific data collection and paradigm-specific model training. In the area of BCIs, 
Large Brainwave Foundation Models (LBMs) have emerged over the past few years [\citet{LaBraM, NeuroGPT}]. These models have struggled to fully capture the rich information content of neural oscillations, a limitation that fundamentally constrains their performance and generalizability across diverse BCI tasks. Brain oscillations, the rhythmic electrical activity that forms the basis of EEG signals, are characterized by both amplitude and phase components. While amplitude reflects the power of neural activity, phase information encodes the precise timing of neural events and the synchronization between brain regions. Existing foundation models for EEG, including the recently introduced LaBraM \cite{LaBraM}, have made significant improvements by leveraging learning techniques from language and vision domains. These approaches typically apply vector-quantized \citet{esser2020taming} neural spectrum prediction to convert continuous EEG signals into discrete neural codes. However, due to the way these models are trained, they suffer from a critical mathematical limitation: \textbf{they fail to properly represent the circular topology of phase information.} 

In this work, we introduce LaBraM++, an enhanced LBM that addresses this fundamental challenge through principled improvements grounded in signal processing theory. Our key insight is that phase information must be represented in a topologically consistent manner that preserves the circular nature of oscillatory dynamics. We accomplish this through \textbf{a mathematically optimal sine/cosine phase representation that maintains the full information content of neural oscillations} while enabling stable gradient-based learning. This approach creates smooth and continuous optimization that facilitates more effective model convergence and leads to representations that better capture the true structure of neural activity.
Beyond addressing the phase representation challenge, LaBraM++ incorporates several complementary elements compared to the original architecture: (1) the application of Common Average Reference (CAR) and Z-scoring to isolate the most relevant neural activity by reducing common noise shared across channels (2) improved temporal and spatial embeddings that better handle the heterogeneity of EEG recording configurations and (3) a redesigned training procedure that enhances the model’s ability to capture meaningful neural patterns.

\section{Background}

LaBraM \citet{LaBraM} is a recently introduced unified EEG foundation model. Inspired by VQ-GAN \citet{esser2020taming}, LaBraM's tokenizer is built on the vector-quantized neural spectrum prediction, trained to predict the Fourier spectrum of EEG signals. The encoding part is based on the neural transformer which receives patches of segmented EEG channels as input. Temporal convolution layers extract temporal features that are then combined with temporal and spatial embeddings and passed through a series of transformer layers \citet{vaswani2023attentionneed}, with the incorporated modification as described in \citet{dehghani23a}. The resulting patch representations are quantized into the neural codebook $\mathcal{V}$. These tokens are then passed into the decoding part (a shallow network) for reconstruction. The target objective is the reconstruction of the Fourier amplitude ($A$) and phase ($\phi$) of the EEG patches by minimizing the total loss:

    \begin{equation}
    \mathcal{L}_T = \sum_{i} 
    \underbrace{\left\| \hat{A_i} - A_i \right\|_2^2}_{\textcolor{purple}{\text{Amplitude loss - $\mathcal{L}_A$}}} + 
    \underbrace{\left\| \hat{\phi_i} - \phi_i \right\|_2^2}_{\textcolor{blue}{\text{Phase loss - $\mathcal{L}_\phi$}}} + 
    \mathcal{L}_Q
    \label{eq:labram_loss}
    \end{equation}

    where $\hat{A}$ and $\hat{\phi}$ are the reconstructed amplitude and phase  (direct amplitude $\mathcal{L}_A$ and direct phase $\mathcal{L}_\phi$ loss functions respectively) and $L_Q$ is the quantization loss as described in \citet{esser2020taming}. While we agree with LaBraM’s core idea of encoding diverse neural activities into discrete neural tokens within a codebook $\mathcal{V}$ and using this codebook to train a large LBM, we identify several mathematical flaws in the codebook’s training process that we aim to address in this work.

\section{Model}

The proposed LaBraM++ model builds upon the original LaBraM introduced by \citet{LaBraM}. The primary contribution of this work is the development and training of an \textbf{improved tokenizer} with enhanced signal reconstruction capabilities. This improvement facilitates the training of a \textbf{more effective} LBM, resulting in superior performance across a range of tasks. Our key insight is that phase information (closely related to brain oscillations) should be represented in a way that respects the inherent circular structure of oscillatory dynamics. To achieve this, we adopt a sine/cosine phase encoding scheme that preserves the full informational content of neural oscillations while supporting stable gradient-based optimization.

% \subsection{Modified Tokenizer}

\begin{figure}[!h]
    \begin{center}
    \includegraphics[width=\linewidth]{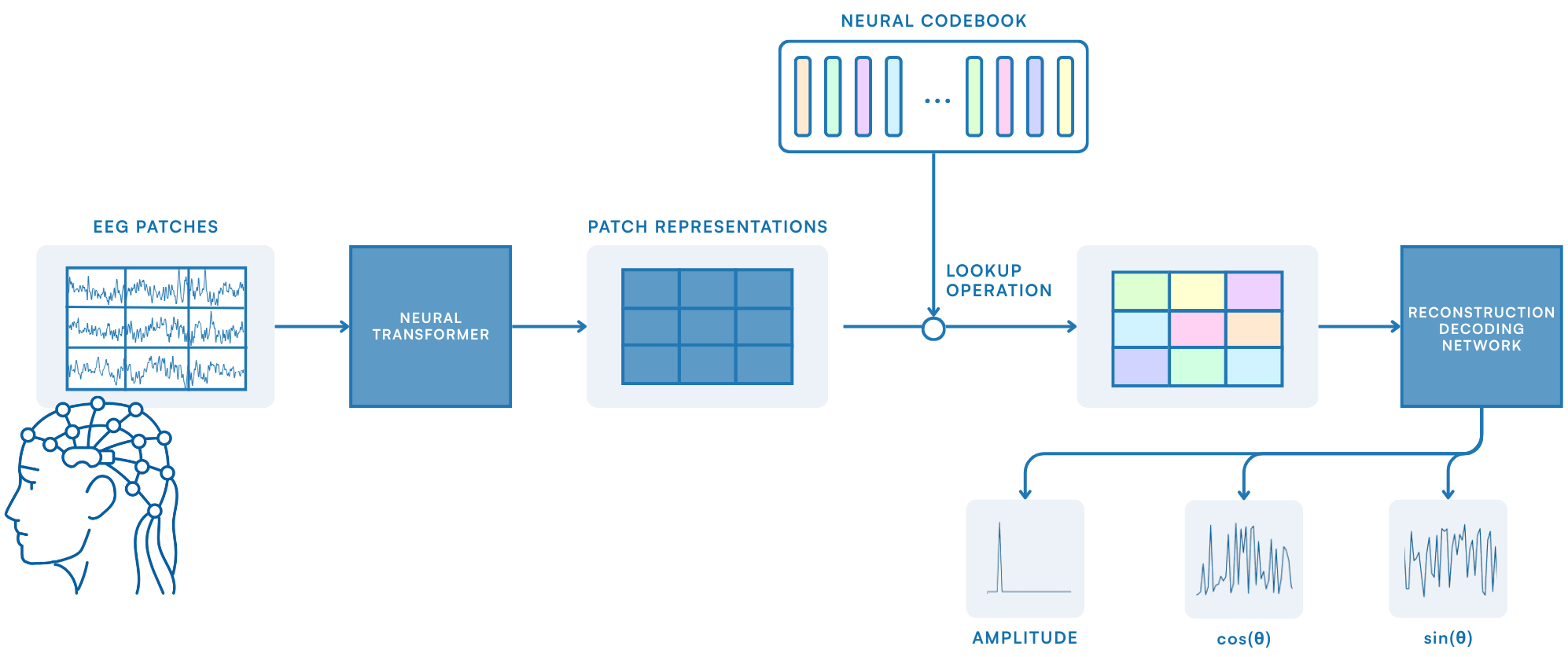}
    \end{center}
    \caption{Illustration of LaBraM++ modified tokenizer. This tokenizer discretizes EEG signals into neural tokens by reconstructing the Fourier amplitude, the sine and the cosine of the Fourier phase.
    }
\end{figure}

\subsection{Description}

Let $X \in \mathbb{R}^{C \times T}$ denote the input EEG signal, where $T$ is the number of time points and $C$ is the number of electrodes. In both the original LaBraM architecture and inherently in LaBraM++, the input EEG is first segmented into patches along the temporal dimension for each channel. Each patch is then passed through the neural transformer (as described in Section 2), which computes a patch representation $\mathbf{p} \in \mathbb{R}^{D}$, where $D$ is the embedding dimension. As described previously, the tokenizer aims to learn a codebook of neural tokens, denoted as $\mathcal{V}$:

\begin{equation}
     \mathcal{V}  = \{\mathbf{v}_{i} | i= 1,...,K\} \in \mathbb{R}^{K \times D}
\end{equation}

where $K$ is the number of discrete neural tokens and $D$ is the dimensionality of each token. A quantizer maps each patch representation $\mathbf{p}$ to its corresponding neural token $\mathbf{v} \in \mathcal{V}$. Specifically, for each patch, the quantizer selects the nearest neighbor $\mathbf{z}$ from the codebook:

\begin{equation}
    \mathbf{z} = \arg\min \left\| l_{2}(\mathbf{p}) - l_{2}(\mathbf{v}) \right\|
\end{equation}

where $l_{2}$ represents normalization.

\textbf{Phase Information Issue:} In the LaBraM architecture, the neural codebook $\mathcal{V}$ is trained via the reconstruction of the Fourier amplitude ($A$) and phase ($\phi$) of the EEG patches by minimizing the total loss $\mathcal{L}_T$, described in equation \ref{eq:labram_loss}, which consists of the direct amplitude $\mathcal{L}_A$ and direct phase $\mathcal{L}_\phi$ loss function terms. However, $\mathcal{L}_\phi$ between actual $\phi$ and reconstructed $\hat{\phi}$ angles is not an ideal loss function due to the periodic nature of the Fourier phase. This can be seen most severely at the boundaries of the phase domain ($\pm \pi$), where similar physiological states produce drastically different representations, causing severe information loss (Section \ref{apx:theory}).

\textbf{Sine/Cosine Phase Loss Function:} To address this issue, we modify the total loss function $\mathcal{L}_T$, described in equation \ref{eq:labram_loss}, by replacing $\mathcal{L}_\phi$ with the sine/cosine phase loss functions ($\mathcal{L}_{sin}$ / $\mathcal{L}_{cos}$) between actual $\phi$ and reconstructed $\hat{\phi}$ angles. In doing so, we preserve the cyclical nature of the phase since both $\cos$ and $\sin$ are smooth and continuous functions across the whole range of angles and therefore more suitable as terms in the total reconstruction loss function $\mathcal{L}_T$. This enables the modified tokenizer to interpolate and generalize more easily. Mathematically, the modified total loss $\mathcal{L}_T$ can be written as: 

\begin{equation}
    \mathcal{L}_T = \sum_{i} 
    \underbrace{\left\| \hat{A_i} - A_i \right\|_2^2}_{\textcolor{purple}{\text{Amplitude loss - $\mathcal{L}_A$}}} + \underbrace{\left\| \sin{\hat{\phi_i}} - \sin{\phi_i} \right\|_2^2}_{\textcolor{olive}{\text{Sine Phase loss - $\mathcal{L}_{sin}$}}} + \underbrace{\left\| \cos{\hat{\phi_i}} - \cos{\phi_i} \right\|_2^2}_{\textcolor{orange}{\text{Cosine Phase loss - $\mathcal{L}_{cos}$}}} + 
    \mathcal{L}_Q
\label{eq:labram_plus_loss}
\end{equation}

where $\hat{A}$, $\sin{\hat{\phi}}$ and $\cos{\hat{\phi}}$ are the reconstructed amplitude and phase sin/cos values respectively, and $L_Q$ is the quantization loss as described in \citet{esser2020taming}. The phases are essentially represented as unit vectors in 2D space, making the errors between them more geometrically meaningful (Section \ref{apx:theory}).

\subsection{Theoretical Basis}
\label{apx:theory}

Although replacing $\mathcal{L}_\phi$ with $\mathcal{L}_{sin}$ and $\mathcal{L}_{cos}$ may initially appear insignificant, this modification is grounded in signal processing theory and offers well-established benefits. Specifically, it contributes to preserving the full informational content of neural oscillations, improving the network's signal reconstruction capabilities and its stable gradient-based optimization.

\textbf{Limitations of Direct Phase Representation:} The direct phase loss function (used in LaBraM) $\mathcal{L}_\phi = ||\hat{\phi}-\phi||^{2}$ between actual $\phi$ and reconstructed $\hat{\phi}$ angles is discontinuous at the boundaries of the phase domain $[-\pi, \pi]$, creating optimization challenges. Consider the case where the predicted phase is $\hat{\phi} = \pi - \epsilon$ and the actual phase is $\phi = -\pi + \epsilon$, with $\epsilon > 0$ small. Using the direct phase loss function: 

\begin{equation}
    \mathcal{L}_\phi = ||\hat{\phi}-\phi||^{2} = ||2\pi - 2\epsilon||^{2} = (2\pi - 2\epsilon)^{2} 
\end{equation}

As $\epsilon \rightarrow 0$, this approaches $4\pi^{2}$, despite representing angles that are close. Therefore, a small phase change, for example of just 2 degrees between $179^{\circ}$ and $-179^{\circ}$, can lead to a huge discontinuity shift. From a machine learning perspective, this problem creates fundamental optimization challenges. The discontinuity at phase boundaries leads to unstable gradients, poor convergence properties and ultimately suboptimal representation learning. This can make optimization unstable and lead to poor signal reconstruction. 

\textbf{Impact on Gradient-Based Optimization:} Using the direct phase loss function $\mathcal{L}_\phi = ||\hat{\phi}-\phi||^{2}$ between actual $\phi$ and reconstructed $\hat{\phi}$ angles leads to unstable gradients and poor optimization properties. For gradient-based optimization, we need to compute:
\begin{equation}
\frac{\partial L_{{\phi}}( \hat{\phi}, \phi)}{\partial \hat{\phi}} = 2(\hat{\phi} - \phi)
\end{equation}

Consider the case where the predicted phase is $\hat{\phi} = \pi - \epsilon$ and the actual phase is $\phi = -\pi + \epsilon$, with $\epsilon > 0$ small. The gradient becomes:

\begin{align}
\frac{\partial L_{{\phi}}( \hat{\phi}, \phi)}{\partial \hat{\phi}} &= 2((\pi - \epsilon) - (-\pi + \epsilon))\\
&= 2(2\pi - 2\epsilon)\\
&= 4\pi - 4\epsilon
\end{align}

As $\epsilon \to 0$, the magnitude of this gradient approaches $4\pi$, which is significantly larger than the gradients elsewhere in the phase domain. The large disparity in gradient magnitudes introduces several significant challenges during optimization. Abrupt changes in gradient values can result in numerical instability. In addition, gradient explosions can take place since excessively large gradients near the boundaries can cause disproportionately large optimization steps, potentially overshooting the optimal solution. In other words, these insights mean that neural networks trained with the direct phase loss function will struggle to learn accurate representations of phases near the $\pm\pi$ boundaries. This mathematical inconsistency creates a representation bottleneck that constrains model performance regardless of architecture size or training data volume.

\textbf{Benefits of Sine/Cosine Phase Representation:} We define the sine/cosine representation of phase $\phi$ as $r(\phi) = (\sin(\phi), \cos(\phi)) \in \mathbb{R}^2$, mapping phase to a point on the unit circle. The sine/cosine phase loss function:

\begin{equation}
    \mathcal{L}_{r(\phi)} = \textcolor{olive}{\mathcal{L}_{sin}} + \textcolor{orange}{\mathcal{L}_{cos}} = ||  \sin(\hat{\phi})- \sin(\phi)||^2 + ||\cos(\hat{\phi}) - \cos(\phi)||^2 = 2 - 2\cos( \hat{\phi} - \phi)
\end{equation}
between actual $\phi$ and predicted angles $\hat{\phi}$ preserves more phase information than direct phase representation, particularly for phases near the $\pm\pi$ boundary.

Consider the case where the predicted phase is $\hat{\phi} = \pi - \epsilon$ and the actual phase is $\phi = -\pi + \epsilon$, with $\epsilon > 0$ small. Using the sine/cosine phase loss function: 

\begin{equation}
    \mathcal{L}_{r(\phi)} = 2 - 2\cos( \hat{\phi} - \phi) = 2 - 2\cos(2\epsilon) \approx 2 - 2(1 - 2\epsilon^2) = 4\epsilon^2
\end{equation}

As $\epsilon \rightarrow 0$, this approaches 0. In the previous example, a phase change between $179^{\circ}$ and $-179^{\circ}$ is now translated to a small smooth difference since $cos(179^{\circ}) \simeq cos(-179^{\circ})$. 

More general, for any phases $\phi_1$ and $\phi_2$, we can show:
\begin{align}
L_{r(\phi)} = \textcolor{olive}{\mathcal{L}_{sin}} + \textcolor{orange}{\mathcal{L}_{cos}} &= 2 - 2\cos(\phi_1 - \phi_2)\\
&= 4\sin^2\left(\frac{\phi_1 - \phi_2}{2}\right)
\end{align}

\begin{tcolorbox}
[colback=blue!5!white,colframe=cyan!75!black]
  \centering
    $L_{r(\phi)}$ is proportional to the squared chord length on the unit circle between points at angles $\phi_1$ and $\phi_2$.  
\end{tcolorbox}
\vskip 0.2in

\subsection{Input Patches}

Although this significant modification in the tokenizer's loss function can lead to smoother training and better reconstruction capabilities, we wanted to further enhance the neural codebook $\mathcal{V}$ by removing certain variances from the EEG signal, encouraging the learning of meaningful neural tokens. Therefore, we perform ablation studies where we apply Common Average Reference (CAR) and Z-scoring to each input EEG patch. CAR reduces common noise shared across channels by subtracting the average activity while Z-scoring standardizes the signal to each patch by removing differences in amplitudes across channels.

\section{Experiments}
\subsection{Implementation Details}
\subsubsection{Data Preprocessing}
\label{sec:preprocess}

For a fair comparison, we aimed to pre-train our models with the same datasets used in the original LaBraM. While we have included the majority of the publicly available datasets, some could not be sourced, including the authors' self-collected dataset. To compensate for this deficit, we also include $\sim$235 hours of our own self-collected motor data \footnote{EEG Data Collection taken place at Imperial College London : https://mybraincommands.doc.ic.ac.uk}.
The full breakdown of included and excluded data is given in Appendix \ref{apx:dataset}.

To preprocess the data, bandpass filter of 0.5-44.5Hz was applied. By choosing this lowpass frequency we also made any notch filters at 50Hz, 60Hz and harmonics redundant, since the powerline noise at these frequencies was already removed. All signals were resampled to 200Hz to match the original implementation. 

We also made adjustments to the structure of the input samples. The original formulation has each sample $\mathbf{x} \in \mathbb{R}^{C_d\times W}$ where $C_d$ is the number of channels for the pre-training dataset $d$. $W$ is defined as $W = t \times w$ where $t$ is the time windows and $w$ the window's length. The $w$ is fixed at 200, i.e. one second (given 200Hz sampling frequency). The value of $t$ is chosen manually for each dataset $d$ based on the $C_d$ such that the total number of patches $P$ $\approx 256$. For example, a $t$ of 8 seconds would be chosen for a dataset with 32 electrodes.

To avoid this manual selection, during model pre-training we set the input size to $P = 256$ patches of length $w=200$, but the number of channels is not fixed. To implement this, we modified the data retrieval process used during pre-training. Rather than iterating through every possible sample in the corpus, we indexed our pre-training database by the individual EEG recording files. From within a given recording, we selected a random trial and use data from a randomly selected subset of electrodes. We then took a random time window from the length of the trial, resulting in the final sample $\mathbf{x} \in \mathbb{R}^{P \times w}$. This randomized retrieval process also allowed more flexibility for pre-training datasets where $C_d$ is not fixed and the number of good quality channels differs between recordings.

\subsubsection{Temporal and Spatial Embeddings}
\label{sec:embeddings}

Since the reconstruction of EEG signal takes place in a per-patch level, the vital temporal and spatial information is incorporated through the use of trainable embedding parameters $TE$ and $SE$, each with embedding dimension $D$. 

Our modifications to pre-training data as detailed in Section \ref{sec:preprocess} require changes to these embeddings compared to the original architecture. Firstly, the embedding list $SE = \{se_1,...,se_{|\mathcal{C}|}\}$ is indexed by the positions of the patches' electrodes within a list of all electrodes in the entire pre-training database $\mathcal{C}$. The length of SE is therefore determined by the length of this global electrode list $|\mathcal{C}|$: for our corpus the total is 104.

Secondly, $TE$ was originally defined as $\{te_1,...,te_{t_{max}}\}$, of which $\{te_1,...,te_{t}\}$ would be used for each sample. This requires setting the hyperparameter $t$, specifying the number of time windows in the sample, as well as the hyperparameter $t_{max}$ denoting its maximum possible value. Our new structure does not need these hand selected parameters because we know each sample, regardless of its source dataset, will consist of $P$ patches of length $w$. We therefore define $TE = \{te_1,...,te_P\}$. By default we use $\{te_1,...,te_{t_{n}}\}$ where $t_{n}$ is the sample's number of patches in the time dimension. However, we also investigate the effect of "right aligning" the time embedding by selecting $\{te_{P-t_{n}},...,te_{P}\}$.

\subsubsection{Hyperparameters}
\label{sec:implementation}

LaBraM++ was trained using the same settings as the original LaBraM architecture (a detailed list of the training settings is provided in Appendix \ref{apx:hypers}). The modified tokenizer was trained for 100 epochs with a neural codebook defined as $\mathcal{V} \in \mathbb{R}^{8192 \times 64}$. Due to computational resource limitations, the base architecture of LaBraM was utilized for the core foundation model, which was trained for 50 epochs. Both models were trained on the datasets described in Appendix \ref{apx:dataset}.

 \subsection{Comparison with LaBraM}

The first step involves comparing LaBraM++ against the original LaBraM \cite{LaBraM}. Since a self-collected dataset has been used, it is not possible to use the exact same datasets as in \citet{LaBraM} for model pre-training. Therefore, for fair comparison we pre-trained both architectures LaBraM++ and LaBraM using the same datasets, pre-processing steps and implementation settings as described in sections \ref{sec:preprocess}, \ref{sec:embeddings}  and \ref{sec:implementation}. 

\subsubsection{Signal Reconstruction}

To qualitatively assess the reconstruction capabilities of LaBraM++’s tokenizer, we visualized the reconstructed EEG signals. As illustrated in the figure below, LaBraM++'s codebook effectively captures the overall trend and low-frequency components of the input signals. However, it shows limitations in accurately reconstructing higher-frequency details, suggesting room for improvement.

\begin{figure}[!h]
  \centering
  \includegraphics[width=\textwidth]{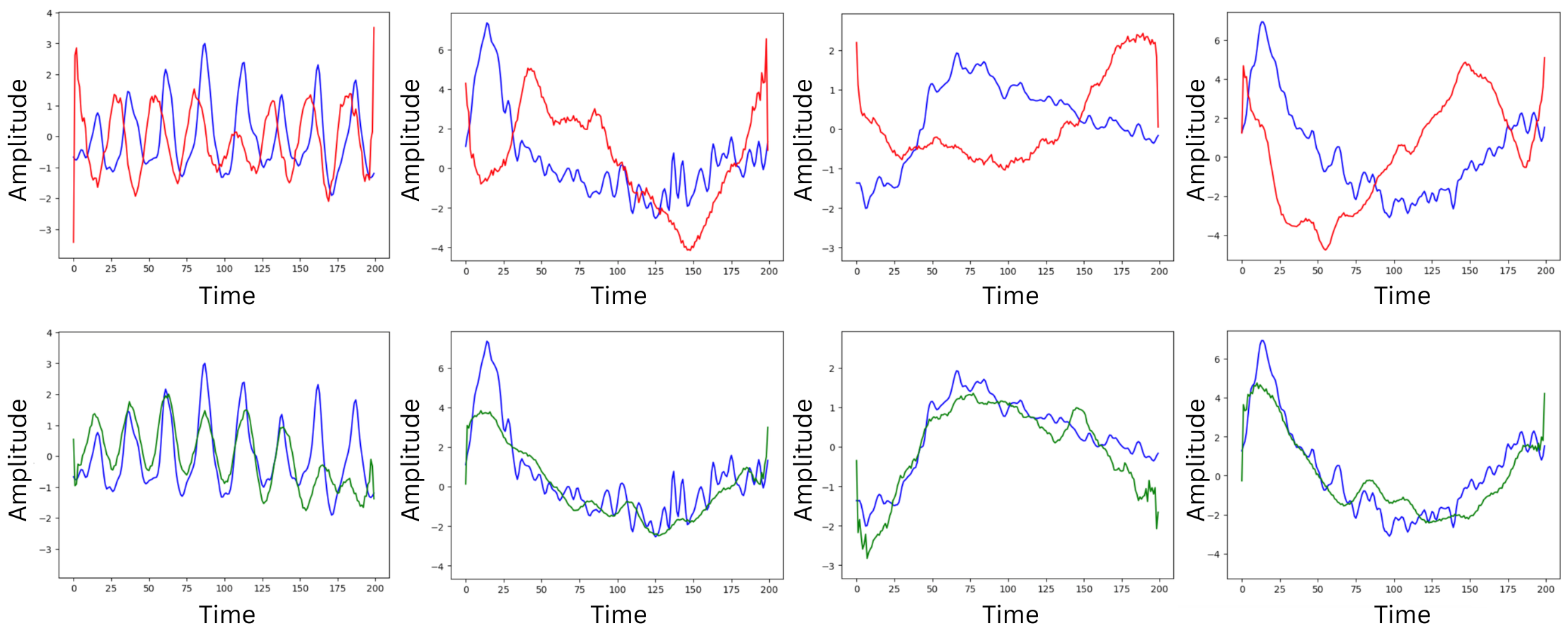}
  \caption{Reconstructed EEG signals from LaBraM (\textcolor{red}{red}) and LaBraM++ (\textcolor{green}{green}) tokenizers. Results show improved reconstructions after applying our modifications. Blue lines denote the input EEG signal.}
  \label{fig:reconstructions}
\end{figure}

\subsubsection{Foundation Model Training Loss}

\begin{figure}[!h]
    \begin{center}
    \includegraphics[width=\linewidth]{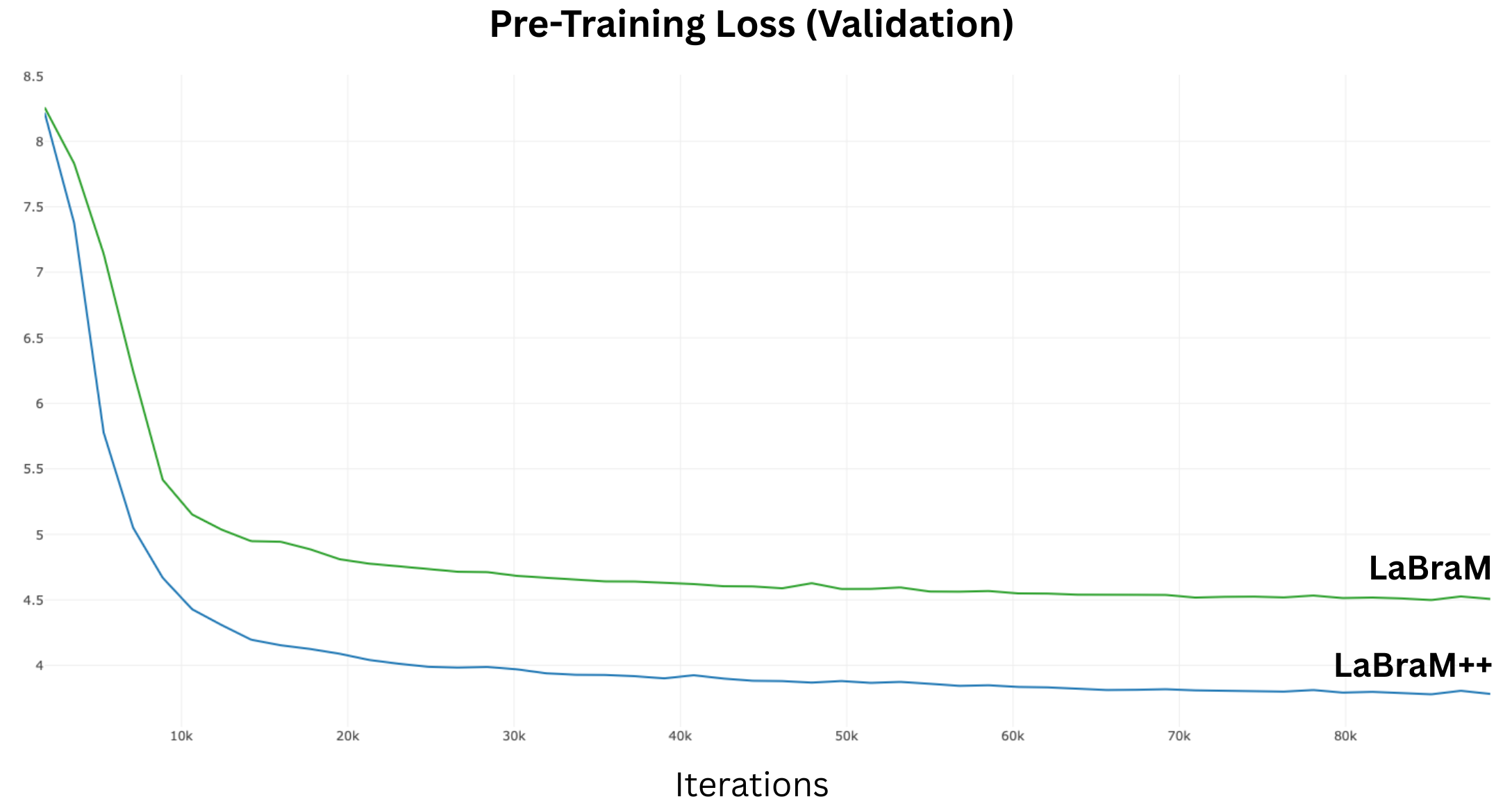}
    \end{center}
    \caption{The pre-training loss curve for LaBraM and LaBraM++}
    \label{fig:curves}
\end{figure}

Figure \ref{fig:curves} presents the pre-training loss curves for both LaBraM and LaBraM++. As shown, LaBraM++ achieves a significantly lower training loss. Although both models share the same architecture (LaBraM base model), these results suggest that the modified tokenizer contributes to a more effective training process for the foundation model.

\subsubsection{Fine-Tuning Performance}
\label{sec:fin}

The next step is to compare the classification performances of these two models. We evaluated the performance of LaBraM++ against the original LaBraM architecture on full finetuning (training of the LBM and the added classification layer). Both models were evaluated in downstream classification tasks for the following four EEG datasets as described in the benchmark \cite{lee2025assessing}.
% Motor paradigm in High Gamma \cite{HighGamma},  a Working Memory dataset \cite{Pavlov22}, Physionet's sleep staging dataset, Sleep-EDF \cite{SleepEDF} and Eyes Open vs Closed classification on the Physionet Motor dataset \cite{physionetmi}. 
These tasks were selected to capture a diverse range of BCI paradigms and the datasets were specifically chosen for their minimal spurious artifacts, reducing the likelihood of spurious performance during training \cite{lee2025assessing}. Each model was trained for 20 epochs (to avoid overfitting) and evaluated using 10-fold subject-independent cross-validation, where samples were split on a subject level such that no subject would be present in both the training and validation sets. To perform the downstream tasks, untrained classification heads were added to the pre-trained LBMs before finetuning. The results of the comparisons are displayed in Table \ref{table: labram_and_plus}.  

\begin{table}[!h]
  \caption{Classification balanced accuracy of LaBraM and LaBraM++ models after finetuning for 20 epochs ($^{*}$ denotes early stopping to avoid overfitting) with 10-fold cross validation. Bold values indicate best performance (per task or overall).}
  \label{ft-table}
  \centering
  \vspace{0.1in}
  \begin{tabular}{l|cccc|c}
    \toprule
    Model & Motor & Memory & Sleep $^{*}$ & Eyes & Mean \\
    \midrule
    LaBraM &
    0.570   & 0.565 & 0.715 & 0.805 & 0.664\\
    LaBraM++ &
    \textbf{0.723}& \textbf{0.584} & \textbf{0.731}& \textbf{0.851} & \textbf{0.722}\\
    \bottomrule
  \end{tabular}
  \label{table: labram_and_plus}
\end{table}

We also performed some ablation studies to introduce variation of LaBraM++ using  Common Average Reference (CAR) and Z-scoring to each patch as well as modification to the temporal embeddings as described in previous sections. The results of these model variations are displayed in Table \ref{table: labram_plus_variations}:

\begin{table}[!h]
  \caption{Classification balanced accuracy of LaBraM++ variations after finetuning for 20 epochs ($^{*}$ denotes early stopping to avoid overfitting) with 10-fold cross validation. RA denotes Time-Embedding Right-Alignment and CAR denotes Common Average Reference. Bold and underlined values indicate best performance and next-best performance respectively (per task or overall).}
  \label{ft-table-2}
  \vspace{0.1in}
  \centering
  \begin{tabular}{l|cccc|c}
    \toprule
    Model & Motor & Memory & Sleep$^{*}$  & Eyes & Mean \\
    \midrule
    LaBraM++ &
    \underline{0.723} & 0.584 & \textbf{0.731} & 0.851 & \underline{0.722} \\
    \midrule
    LaBraM++ with CAR &
    \underline{0.723} & 0.580 & \underline{0.699} & 0.847 & 0.712 \\
    LaBraM++ with CAR and Z-scoring & 
    0.715 & \underline{0.605} & 0.667 & \textbf{0.876} & 0.716 \\
    LaBraM++ with CAR, Z-scoring and RA &
    \textbf{0.741} &  \textbf{0.616} & 0.673  & \underline{0.873} & \textbf{0.726} \\
    \bottomrule
  \end{tabular}
  \label{table: labram_plus_variations}
\end{table}

As shown in Tables \ref{table: labram_and_plus} and \ref{table: labram_plus_variations}, LaBraM++ substantially outperforms the original LaBraM model, achieving a 6\% improvement in overall performance.  LaBraM++ also demonstrates a significantly lower training loss, indicating more effective model optimization. Importantly, the modified tokenizer (the main contribution of this work) demonstrates great reconstruction capabilities compared to the original LaBraM. Taken together, these results provide strong evidence that LaBraM++ constitutes a meaningful advancement over LaBraM, offering improved performance without compromising on model capacity or training efficiency.

\subsection{Comparison with State-of-the-Art}

We evaluated the performance of fine-tuned LaBraM++ in comparison to other fine-tuned LBMs, namely variations of NeuroGPT \cite{NeuroGPT}, BIOT \cite{yang2023biot}, EEGPT \cite{wang2024eegpt} and CBraMod \cite{wang2025cbramod}. By using the same benchmarking as described in section \ref{sec:fin} and originally introduced in \citet{lee2025assessing}, we perform the same full finetuning and use the same 10-fold subject-independent cross-validation for the evaluation of all models. As shown in Table \ref{table: sota_lbms}, when full-finetuned, LaBraM++ demonstrates competitive performance against other leading finetuned LBM models. This suggests that training LaBraM++ on a broader range of datasets used by existing LBMs could further enhance its performance, reinforcing its potential as a strong and efficient brainwave foundation model.

\vspace{-0.1in}
\begin{table}[!h]
  \caption{Classification balanced accuracy of finetuned foundation models. Each trained/finetuned for 20 epochs ($^{*}$ denotes early stopping to avoid overfitting) with 10 fold cross-validation. Bold and underlined values indicate best performance and next-best performance respectively (per task or overall). RA denotes Time-Embedding Right-Alignment and CAR denotes Common Average Reference.}
  \centering
  \vspace{0.1in}
  \begin{tabulary}{\linewidth}{L|cccc|c}
    \toprule
    Model & Motor & Memory & Sleep & Eyes & Mean \\
    \midrule
    NeuroGPT (full model) & 0.682  & 0.597  & \underline{0.674} & 0.827  & 0.695  \\ 
    NeuroGPT (encoder) & 0.695  & \textbf{0.653}  & 0.667  & 0.838 & 0.713  \\
    CBraMod & 0.614 & 0.574 & 0.635 & 0.839 & 0.666 \\
    % NeuroLM & & & & & \\
    BIOT & 0.443 & 0.510 & - & 0.763 & -\\
    % LaBraM &
    % 0.570   & 0.565 & 0.715 & 0.805 & 0.664 \\
    EEGPT & 0.313 &	0.520 &	0.634	 & 0.797	& 0.566 \\
    \midrule
    LaBraM++ & 
    \underline{0.723} & 0.584 & \textbf{0.731}$^{*}$ & \underline{0.851} & \underline{0.722}   \\
    LaBraM++ with CAR, Z-scoring and RA &
    \textbf{0.741} & \underline{0.616} & 0.673$^{*}$ & \textbf{0.873} & \textbf{0.726} \\
    \bottomrule
  \end{tabulary}
  \label{table: sota_lbms}
\end{table}

\newpage
\section{Discussion}

In this work, we introduce LaBraM++, an LBM with refined components compared to the original LaBraM model introduced in \cite{LaBraM}. By utilizing the benchmarking protocol described in \cite{lee2025assessing}, our findings reveal that LaBraM++ significantly outperforms the original LaBraM architecture when trained on the same datasets, achieving a 6\% improvement in overall performance. When fully fine-tuned, LaBraM++ performs competitively against other pre-trained, open-source LBMs.

In addition, the qualitative assessment of the reconstruction capabilities of LaBraM++’s tokenizer has shown that LaBraM++’s codebook effectively captures the overall trend and low-frequency components of the input signals with limitations in accurately reconstructing higher-frequency details. One promising avenue for enhancing LaBraM++’s performance lies in rethinking the signal reconstruction target. Instead of relying only on Fourier transform, which can miss short-term or localized frequency details, future versions of LaBraM++ could use other additional transformations like scattering transform. Additionally, incorporating training principles from the field of causal reasoning may further boost model performance. As highlighted in a recent work on causal modeling for LBMs \cite{barmpas2024a}, careful masking selection, both temporal and spatial, during the training phase is essential for capturing critical neural activity accurately during
training. 

\section{Conclusion}

In this work, we introduce LaBraM++, a Large Brainwave Foundation Model (LBM) built upon the original LaBraM introduced \citet{LaBraM}. The main contribution of this work is the design and training of an improved tokenizer, which in turn facilitates the development of a more effective LBM that shows improved performance across a range of tasks. Through a series of experiements, we showed that LaBraM++ significantly outperforms the original LaBraM architecture while achieving competitive performance against other pre-trained open-source LBMs. In summary, LaBraM++ marks a step forward in the development of LBMs. Its improvements in performance and training efficiency, along with the potential for further enhancement through other signal representations and causal reasoning frameworks, set a strong foundation for future research in the area of LBMs.

\newpage

\bibliographystyle{nips0}
\bibliography{neurips_2025}

\begin{thebibliography}{41}
\providecommand{\natexlab}[1]{#1}
\expandafter\ifx\csname urlstyle\endcsname\relax
  \providecommand{\doi}[1]{doi:\discretionary{}{}{}#1}\else
  \providecommand{\doi}{doi:\discretionary{}{}{}\begingroup \urlstyle{rm}\Url}\fi

\bibitem[Bakas et~al., 2022]{ourNeurISP2021}
Bakas, S., Ludwig, S., Barmpas, K., Bahri, M., Panagakis, Y., Laskaris, N., Adamos, D.~A. \& Zafeiriou, S. (2022) Team cogitat at neurips 2021: Benchmarks for eeg transfer learning competition.

\bibitem[Barmpas et~al., 2024{\natexlab{a}}]{barmpas2024a}
Barmpas, K., Panagakis, Y., Adamos, D., Laskaris, N. \& Zafeiriou, S. (2024{\natexlab{a}}) A causal perspective in brainwave foundation models.
\newblock In \emph{Causality and Large Models @NeurIPS 2024}.

\bibitem[Barmpas et~al., 2023]{Barmpas_Scattering}
Barmpas, K., Panagakis, Y., Adamos, D.~A., Laskaris, N. \& Zafeiriou, S. (2023) Brainwave-scattering net: a lightweight network for eeg-based motor imagery recognition.
\newblock \emph{Journal of Neural Engineering}, 20(5):056014.
\newblock ISSN 1741-2552.

\bibitem[Barmpas et~al., 2024{\natexlab{b}}]{Barmpas_2024_causality_jne}
Barmpas, K., Panagakis, Y., Zoumpourlis, G., Adamos, D.~A., Laskaris, N. \& Zafeiriou, S. (2024{\natexlab{b}}) A causal perspective on brainwave modeling for brain–computer interfaces.
\newblock \emph{Journal of Neural Engineering}, 21(3):036001.
\newblock \doi{10.1088/1741-2552/ad3eb5}.

\bibitem[Bashashati et~al., 2007]{SignalProcessingMethods1}
Bashashati, A., Fatourechi, M., Ward, R.~K. \& Birch, G.~E. (2007) A survey of signal processing algorithms in brain{\textendash}computer interfaces based on electrical brain signals.
\newblock \emph{Journal of Neural Engineering}, 4(2):R32--R57.
\newblock \doi{10.1088/1741-2560/4/2/r03}.

\bibitem[Blankertz et~al., 2007]{bci_iv1}
Blankertz, B., Dornhege, G., Krauledat, M., Müller, K.-R. \& Curio, G. (2007) The non-invasive berlin brain–computer interface: Fast acquisition of effective performance in untrained subjects.
\newblock \emph{NeuroImage}, 37(2):539--550.
\newblock ISSN 1053-8119.
\newblock \doi{https://doi.org/10.1016/j.neuroimage.2007.01.051}.

\bibitem[Brown et~al., 2020]{brown2020language}
Brown, T.~B., Mann, B., Ryder, N., Subbiah, M., Kaplan, J., Dhariwal, P., Neelakantan, A., Shyam, P., Sastry, G., Askell, A., Agarwal, S., Herbert-Voss, A., Krueger, G., Henighan, T., Child, R., Ramesh, A., Ziegler, D.~M., Wu, J., Winter, C., Hesse, C., Chen, M., Sigler, E., Litwin, M., Gray, S., Chess, B., Clark, J., Berner, C., McCandlish, S., Radford, A., Sutskever, I. \& Amodei, D. (2020) Language models are few-shot learners.

\bibitem[Buckwalter et~al., 2021]{TUAR}
Buckwalter, G., Chhin, S., Rahman, S., Obeid, I. \& Picone, J. (2021) Recent advances in the tuh eeg corpus: Improving the interrater agreement for artifacts and epileptiform events.
\newblock In \emph{2021 IEEE Signal Processing in Medicine and Biology Symposium (SPMB)}, pp. 1--3.
\newblock \doi{10.1109/SPMB52430.2021.9672302}.

\bibitem[Cui et~al., 2024]{NeuroGPT}
Cui, W., Jeong, W., Thölke, P., Medani, T., Jerbi, K., Joshi, A.~A. \& Leahy, R.~M. (2024) Neuro-gpt: Towards a foundation model for eeg.

\bibitem[Dehghani et~al., 2023]{dehghani23a}
Dehghani, M., Djolonga, J., Mustafa, B., Padlewski, P., Heek, J., Gilmer, J., Steiner, A., Caron, M., Geirhos, R., Alabdulmohsin, I., Jenatton, R., Beyer, L., Tschannen, M., Arnab, A., Wang, X., Riquelme, C., Minderer, M., Puigcerver, J., Evci, U., Kumar, M., van Steenkiste, S., Elsayed, G.~F., Mahendran, A., Yu, F., Oliver, A., Huot, F., Bastings, J., Collier, M.~P., Gritsenko, A., Birodkar, V., Vasconcelos, C., Tay, Y., Mensink, T., Kolesnikov, A., Pavetić, F., Tran, D., Kipf, T., Lučić, M., Zhai, X., Keysers, D., Harmsen, J. \& Houlsby, N. (2023) Scaling vision transformers to 22 billion parameters.

\bibitem[Detti et~al., 2020]{sienascalp}
Detti, P., Vatti, G. \& Zabalo Manrqiue~de Lara, G. (2020) Eeg synchronization analysis for seizure prediction: A study on data of noninvasive recordings.
\newblock \emph{Processes}, 8:846.
\newblock \doi{10.3390/pr8070846}.

\bibitem[Esser et~al., 2020]{esser2020taming}
Esser, P., Rombach, R. \& Ommer, B. (2020) Taming transformers for high-resolution image synthesis.

\bibitem[Handy, 2009]{SignalProcessingMethods2}
Handy, T.~C. (2009) \emph{Brain Signal Analysis: Advances in Neuroelectric and Neuromagnetic Methods}.
\newblock The MIT Press.
\newblock ISBN 9780262013086.
\newblock \doi{10.7551/mitpress/9780262013086.001.0001}.

\bibitem[Jiang et~al., 2024]{LaBraM}
Jiang, W., Zhao, L. \& liang Lu, B. (2024) Large brain model for learning generic representations with tremendous {EEG} data in {BCI}.
\newblock In \emph{The Twelfth International Conference on Learning Representations}.

\bibitem[Korczowski et~al., 2019]{bi2015a}
Korczowski, L., Cederhout, M., Andreev, A., Cattan, G., Rodrigues, P., Gautheret, V. \& Congedo, M. (2019) Brain invaders calibration-less p300-based bci with modulation of flash duration dataset (bi2015a).
\newblock \doi{10.5281/zenodo.3266930}.

\bibitem[Lawhern et~al., 2018]{EEGNet}
Lawhern, V.~J., Solon, A.~J., Waytowich, N.~R., Gordon, S.~M., Hung, C.~P. \& Lance, B.~J. (2018) Eegnet: a compact convolutional neural network for eeg-based brain–computer interfaces.
\newblock \emph{Journal of Neural Engineering}, 15(5):056013.
\newblock ISSN 1741-2552.
\newblock \doi{10.1088/1741-2552/aace8c}.

\bibitem[Lee et~al., 2025]{lee2025assessing}
Lee, N., Bakas, S., Barmpas, K., Panagakis, Y., Adamos, D., Laskaris, N. \& Zafeiriou, S. (2025) Assessing the capabilities of large brainwave foundation models.
\newblock In \emph{Workshop on Spurious Correlation and Shortcut Learning: Foundations and Solutions}.

\bibitem[Luciw et~al., 2014]{grasplift}
Luciw, M., Jarocka, E. \& Edin, B. (2014) Multi-channel eeg recordings during 3,936 grasp and lift trials with varying weight and friction.
\newblock \emph{Scientific data}, 1:140047.
\newblock \doi{10.1038/sdata.2014.47}.

\bibitem[Margaux et~al., 2012]{inriaBCI}
Margaux, P., Maby, E., Daligault, S., Bertrand, O. \& Mattout, J. (2012) Objective and subjective evaluation of online error correction during p300-based spelling.
\newblock \emph{Advances in Human-Computer Interaction}, 2012.
\newblock \doi{10.1155/2012/578295}.

\bibitem[McFarland et~al., 2006]{Classical}
McFarland, D., Anderson, C., Muller, K.-R., Schlogl, A. \& Krusienski, D. (2006) Bci meeting 2005-workshop on bci signal processing: feature extraction and translation.
\newblock \emph{IEEE Transactions on Neural Systems and Rehabilitation Engineering}, 14(2):135--138.
\newblock \doi{10.1109/TNSRE.2006.875637}.

\bibitem[Mizrahi et~al., 2023]{mizrahi2023m}
Mizrahi, D., Bachmann, R., Kar, O.~F., Yeo, T., Gao, M., Dehghan, A. \& Zamir, A. (2023) 4m: Massively multimodal masked modeling.
\newblock In \emph{Thirty-seventh Conference on Neural Information Processing Systems}.

\bibitem[Nam et~al., 2018]{SignalProcessingMethods4}
Nam, C., Nijholt, A. \& Lotte, F., eds. (2018) \emph{Brain-Computer Interfaces Handbook: Technological and Theoretical Advances}.
\newblock CRC Press (Taylor \& Francis).
\newblock ISBN 9781498773430.

\bibitem[Paraperas~Papantoniou et~al., 2024]{paraperas2024arc2face}
Paraperas~Papantoniou, F., Lattas, A., Moschoglou, S., Deng, J., Kainz, B. \& Zafeiriou, S. (2024) Arc2face: A foundation model for id-consistent human faces.
\newblock In \emph{Proceedings of the European Conference on Computer Vision (ECCV)}.

\bibitem[Rao, 2013]{SignalProcessingMethods3}
Rao, R.~P. (2013) \emph{Brain-Computer Interfacing: An Introduction}.
\newblock USA: Cambridge University Press.
\newblock ISBN 0521769418.

\bibitem[Santamaría-Vázquez et~al., 2020]{EEGInception}
Santamaría-Vázquez, E., Martínez-Cagigal, V., Vaquerizo-Villar, F. \& Hornero, R. (2020) Eeg-inception: A novel deep convolutional neural network for assistive erp-based brain-computer interfaces.
\newblock \emph{IEEE Transactions on Neural Systems and Rehabilitation Engineering}, 28(12):2773--2782.
\newblock \doi{10.1109/TNSRE.2020.3048106}.

\bibitem[Savran et~al., 2006]{emobrain}
Savran, A., Çiftçi, K., Chanel, G., Mota, J., Viet, L., Sankur, B., Akarun, L., Caplier, A. \& Rombaut, M. (2006) Emotion detection in the loop from brain signals and facial images.

\bibitem[Schalk et~al., 2004]{physionetmi}
Schalk, G., McFarland, D.~J., Hinterberger, T., Birbaumer, N. \& Wolpaw, J.~R. (2004) {BCI2000}: a general-purpose brain-computer interface ({BCI}) system.
\newblock \emph{IEEE Trans. Biomed. Eng.}, 51(6):1034--1043.

\bibitem[Shah et~al., 2018]{TUSZ}
Shah, V., von Weltin, E., Lopez, S., McHugh, J.~R., Veloso, L., Golmohammadi, M., Obeid, I. \& Picone, J. (2018) The temple university hospital seizure detection corpus.

\bibitem[Song et~al., 2023]{EEGConformer}
Song, Y., Zheng, Q., Liu, B. \& Gao, X. (2023) Eeg conformer: Convolutional transformer for eeg decoding and visualization.
\newblock \emph{IEEE Transactions on Neural Systems and Rehabilitation Engineering}, 31:710--719.
\newblock \doi{10.1109/TNSRE.2022.3230250}.

\bibitem[Torkamani-Azar et~al., 2019]{spisresting}
Torkamani-Azar, M., Kanik, S.~D., Aydin, S. \& Cetin, M. (2019) Prediction of reaction time and vigilance variability from spatiospectral features of resting-state eeg in a long sustained attention task.

\bibitem[Touvron et~al., 2023]{touvron2023llama}
Touvron, H., Lavril, T., Izacard, G., Martinet, X., Lachaux, M.-A., Lacroix, T., Rozière, B., Goyal, N., Hambro, E., Azhar, F., Rodriguez, A., Joulin, A., Grave, E. \& Lample, G. (2023) Llama: Open and efficient foundation language models.

\bibitem[Trujillo, 2020]{Trujillo2020}
Trujillo, L. (2020) {Raw EEG Data}.
\newblock \doi{10.18738/T8/SS2NHB}.

\bibitem[Trujillo et~al., 2017]{Trujillo2017}
Trujillo, L.~T., Stanfield, C.~T. \& Vela, R.~D. (2017) The effect of electroencephalogram (eeg) reference choice on information-theoretic measures of the complexity and integration of eeg signals.
\newblock \emph{Frontiers in Neuroscience}, Volume 11 - 2017.
\newblock ISSN 1662-453X.
\newblock \doi{10.3389/fnins.2017.00425}.

\bibitem[Vaswani et~al., 2023]{vaswani2023attentionneed}
Vaswani, A., Shazeer, N., Parmar, N., Uszkoreit, J., Jones, L., Gomez, A.~N., Kaiser, L. \& Polosukhin, I. (2023) Attention is all you need.

\bibitem[Veloso et~al., 2017]{TUEP}
Veloso, L., McHugh, J., von Weltin, E., Lopez, S., Obeid, I. \& Picone, J. (2017) Big data resources for eegs: Enabling deep learning research.
\newblock In \emph{2017 IEEE Signal Processing in Medicine and Biology Symposium (SPMB)}, pp. 1--3.
\newblock \doi{10.1109/SPMB.2017.8257044}.

\bibitem[von Weltin et~al., 2017]{TUSL}
von Weltin, E., Ahsan, T., Shah, V., Jamshed, D., Golmohammadi, M., Obeid, I. \& Picone, J. (2017) Electroencephalographic slowing: A primary source of error in automatic seizure detection.
\newblock \doi{10.1109/SPMB.2017.8257018}.

\bibitem[Wang et~al., 2024]{wang2024eegpt}
Wang, G., Liu, W., He, Y., Xu, C., Ma, L. \& Li, H. (2024) {EEGPT}: Pretrained transformer for universal and reliable representation of {EEG} signals.
\newblock In \emph{The Thirty-eighth Annual Conference on Neural Information Processing Systems}.

\bibitem[Wang et~al., 2025]{wang2025cbramod}
Wang, J., Zhao, S., Luo, Z., Zhou, Y., Jiang, H., Li, S., Li, T. \& Pan, G. (2025) {CB}ramod: A criss-cross brain foundation model for {EEG} decoding.
\newblock In \emph{The Thirteenth International Conference on Learning Representations}.

\bibitem[Wei et~al., 2022]{BEETL}
Wei, X., Faisal, A.~A., Grosse-Wentrup, M., Gramfort, A., Chevallier, S., Jayaram, V., Camille~Jeunet, S.~B., Ludwig, S., Barmpas, K., Bahri, M., Panagakis, Y., Laskaris, N., Adamos, D.~A., Zafeiriou, S., Duong, W.~C., Gordon, S.~M., Lawhern, V.~J., Śliwowski, M., Rouanne, V. \& Tempczyk, P. (2022) 2021 beetl competition: Advancing transfer learning for subject independence \& heterogenous eeg data sets.

\bibitem[Yang et~al., 2023]{yang2023biot}
Yang, C., Westover, M.~B. \& Sun, J. (2023) {BIOT}: Biosignal transformer for cross-data learning in the wild.
\newblock In \emph{Thirty-seventh Conference on Neural Information Processing Systems}.

\bibitem[Zheng \& Lu, 2015]{SEED}
Zheng, W.-L. \& Lu, B.-L. (2015) Investigating critical frequency bands and channels for eeg-based emotion recognition with deep neural networks.
\newblock \emph{IEEE Transactions on Autonomous Mental Development}, 7(3):162--175.
\newblock \doi{10.1109/TAMD.2015.2431497}.

\end{thebibliography}

\newpage
\appendix

\section{Model Configuration and Hyperparameter Settings}
\label{apx:hypers}

This section provides a detailed description of the configuration settings for the temporal encoder module, which is part of both the original and improved versions of the proposed tokenizer. Additionally, it outlines the hyperparameter selections used in training the LaBraM++'s tokenizer, the core foundation model and the fine-tuning procedures for downstream tasks. The experiments were run on Google Cloud using NVIDIA L4 instances with 8 GPUs (16GB memory per GPU) and 384GB of RAM.

\begin{table}[!h]
  \caption{Configuration of temporal encoder module. This same configuration is used in both the tokenizer and core foundation model.}
  \centering
    \begin{tabulary}{\linewidth}{Lccccccc}
    \toprule
    & Layer & Shape & Kernel & Stride & Padding & Norm(N, C) & Activation \\
    \midrule
    \multirow{3}{40pt}{Patch Embedding} & Conv2d & (1, 8) & (1, 15) & (1, 8) & (0, 7) & GroupNorm(4, 8) & GELU \\
    & Conv2d & (8, 8) & (1, 3) & (1, 1) & (0, 1) & GroupNorm(4, 8) & GELU \\
    & Conv2d & (8, 8) & (1, 3) & (1, 1) & (0, 1) & GroupNorm(4, 8) & GELU \\
    \bottomrule
  \end{tabulary}
\end{table}

\begin{table}[!h]
\caption{Hyperparameters for training tokenizer, pre-training core foundation model, and finetuning on downstream tasks.}
\centering
  \begin{tabular}{lccc}
    \toprule
    Hyperparameter & Tokenizer & Pre-training FM & Finetuning \\
    \midrule
    Batch size & 1024 & 512 & 64 \\
    Learning rate scheduler & Cosine & Cosine & Linear \\
    Base learning rate & 5e-5 & 5e-4 & 5e-4 \\
    Min learning rate & 1e-5 & 1e-5 & - \\
    Warmup lr start-end factors & - & - & (0.1,1) \\
    Lr start-end factors & - & - & (1,0.1) \\
    Total epochs & 100 & 50 & 20 \\
    Warmup epochs & 10 & 5 & 4 \\
    Optimizer & AdamW & AdamW & AdamW \\
    Weight decay & 1e-4 & 0.05 & 0.01 \\
    Adam $\beta$ & (0.9, 0.999) & (0.9, 0.999) & (0.9, 0.999) \\   
    Layer lr decay & - & - & 0.975 \\
    \midrule
    Gradient clipping & 3 & - & - \\
    Layer scale init & 0.001 & 0.001 & - \\
    Encoder depth & 12 & 12 & 12 \\
    Decoder depth & 3 & - & - \\
    Hidden dimension & 200 & 200 & 200 \\
    No. Attention heads & 10 & 10 & 10 \\
    MLP hidden dimension & 800 & 800 & 800 \\
    \bottomrule
  \end{tabular}
\end{table}

\newpage
\section{Datasets}
\label{apx:dataset}

For fair comparison, we pre-trained all models in the experimental section of this work using the same following datasets (datasets mentioned in columns Both and LaBraM++ were used). 

\begin{table}[!h]
  \caption{Pre-training data used in the original LaBraM vs LaBraM++}
  \label{dataset-table}
  \centering
  \begin{tabular}{lll}
    \toprule
    LaBraM & Both & LaBraM++ \\
    \midrule
    Emobrain \citet{emobrain} & BCI Competition IV-1 \citet{bci_iv1} & Self-Collected Dataset\\
    SEED \citet{SEED} & Grasp and Lift \citet{grasplift} & \\
    TUSL \citet{TUSL} & Inria BCI Challenge \citet{inriaBCI} & \\
    Self collected \cite{LaBraM} & Physionet MI \cite{physionetmi} & \\
    & Trujillo 2020 \citet{Trujillo2020} & \\
    & Trujillo 2017 \citet{Trujillo2017} &  \\
    & Siena Scalp \citet{sienascalp} & \\
    & SPIS Resting \citet{spisresting} & \\
    & bi2015a \citet{bi2015a} & \\
    & TUAR \citet{TUAR} & \\
    & TUEP \citet{TUEP} & \\
    & TUSZ \cite{TUSZ} & \\
    \bottomrule
  \end{tabular}
\end{table}

\end{document}